# Title

SALIENCE-AFFECTED NEURAL NETWORKS

# Authors

Leendert A Remmelzwaal[1], Jonathan Tapson[1] & George F R Ellis[2]

# Abstract

We present a simple neural network model which combines a locally-connected feedforward structure, as is traditionally used to model inter-neuron connectivity, with a layer of undifferentiated connections which model the diffuse projections from the human limbic system to the cortex. This new layer makes it possible to model global effects such as salience, at the same time as the local network processes task-specific or local information. This simple combination network displays interactions between salience and regular processing which correspond to known effects in the developing brain, such as enhanced learning as a result of heightened affect.

The cortex biases neuronal responses to affect both learning and memory, through the use of *diffuse projections* from the limbic system to the cortex. Standard ANNs do not model this non-local flow of information represented by the ascending systems, which are a significant feature of the structure of the brain, and although they do allow associational learning with multiple-trial, they simply don't provide the capacity for one-time learning.

In this research we model this effect using an artificial neural network (ANN), creating a salience-affected neural network (SANN). We adapt an ANN to embody the capacity to respond to an *input salience signal* and to produce a *reverse salience signal* during testing.

This research demonstrates that input combinations similar to the inputs in the training data sets will produce similar *reverse salience signals* during testing. Furthermore, this research has uncovered a novel method for training ANNs with a single training iteration.

[1] Department of Electrical Engineering, University of Cape Town, Cape Town, South Africa.

[2] Department of Mathematics and Applied Mathematics, University of Cape Town, Cape Town, South Africa.

Correspondence should be addressed to: L.R. (leenremm@gmail.com).

**Glossary**

ANN = artificial neural network
NMF = non-negative matrix factorization
NN = neural network
PCA = principal component analysis
SANN = salience-affected neural network
VQ = vector quantization

**Introduction**

Apart from motor and sensory systems, the human brain has at least two classes of connections relating to the cortex[1]. The first class consists of localized inter-neuronal connections forming layered neural networks, which are modeled by standard artificial neural networks (ANN). Standard ANNs[2] model the way signals flow locally in the columns in the cortex through local connections in those columns, allowing complex processes such as pattern recognition[2] and naming[3] to occur.

Another class consists of diffuse projections from the limbic system to the cortex are known as *monoamine systems*,[4] or *ascending systems*[5]. From their nuclei of origin they send axons up and down the nervous system in a diffused spreading pattern. The effect of the monoamine systems projecting profusely is that each associated neurotransmitter (for example norepinephrine and dopamine) affects large populations of neurons, allowing non-local interactions to occur in the brain. The release of the neurotransmitter affects the probability that neurons in the neighbourhood of value-system axons will fire after receiving glutamatergic input, thus they are an important mechanism effecting neural plasticity. These systems bias neuronal responses affecting both learning and memory by guiding neuronal group selection, and for this reason that they are sometimes termed *value systems*[5,6]. Standard ANNs do not model this non-local flow of information represented by the ascending systems, which are a significant feature of the structure of the brain. The aim of this paper is to model the effects of such connections in a new class of ANNs, which we refer to as *Salience-Affected Neural Networks* (SANNs) because they allow representation of salience effects.

The *salience* of an entity refers to its state or quality of standing out relative to neighboring entities. For example, a salient memory might be one that significantly stands out among others because of its emotional content. Beebe and Lachmann defined the three principles of salience that describe the interaction structures in the first year of life[7]. These are the principles of *ongoing regulations*, *disruption and repair*, and *heightened affective moments*. Ongoing regulation describes the characteristic pattern of repeated interactions, such as a child interacting with their mother or father. This is well represented by standard ANNs, which allow associational learning with multiple-trial (i.e. *ongoing regulations)*. However heightened affective moments are dramatic moments standing out among other events[7], often leading to one-time learning, which is an established psychological phenomenon[8], and has been the subject of discussion in the psychology literature for over 50 years.

While the details of how single-trial training works in the cortex are in dispute, the fact that single-trial training sometimes occurs is well established[9,10]. When it occurs, single-trial learning is often associated with the neurotransmitter dopamine[11]. This is one of the neurotransmitters released throughout the cortex by the ascending systems, and is associated both with the neural coding of basic rewards, and in reinforcement signals defining an agent's goals[12,13]; hence it is known to play an important role in brain functioning. Standard ANNs do not currently model the salience effects of neurotransmitters such as dopamine, and are unable to implement one-time learning; but SANNs do both.

The only other class of ANNs of which we are aware that represent non-local effects are models of the effects of Nitric Oxide (NO), which is a freely diffusing neurotransmitter[14]. A model of gas diffusion is used in a class of ANNs in which units are capable of non-locally modulating the behaviour of other units. These models are called GasNets[14], as they simulate the presence of the gas NO in the environment surrounding the neurons. Like SANNs, this form of modulation allows a kind of plasticity in the network in which the intrinsic properties of units are changing as the network operates[15]. However to the best of our knowledge, GasNets have not been modified to train and test an ANN with specific salience, nor do they have the potential for single-trial learning in ANNs.

In summary, a SANN models the physiological feature of non-local signaling in the cortex via neurotransmitters associated with rewards and reinforcement signals broadcast from the limbic system, and demonstrates that this makes one-time learning possible. Thus it throws light on the relation between these important structural and functional aspects of the human brain, and thereby opens up a new class of ANN that may be useful in computational applications.

In particular an SANN demonstrates that these kinds of networks not only enable one-time learning by laying down memory patterns associated with strong salience signals, but enables recall of those salience signals when the relevant stimulus is encountered at later times; that is, it models both the way memories associated with emotionally-laden events[16] may be embodied in neural networks, and the way those emotional associations can be recalled when the events are remembered.

**Implementation**

In this research, we have used as a platform a generic fully-connected multilayer perceptron (MLP) artificial neural network (ANN). The methods used here should be easily adapted for use with many alternative ANN types. The MLP ANN was selected because it is a well-characterized and common neural network from which many other neural networks are derived.

We define the *salience* of an entity as its state or quality of standing out relative to neighboring entities, *input salience signal* as the additional input signal affecting each node during training, and *reverse salience signal* as the combination of the *nodal reverse salience signals* produced by each node during testing.

To incorporate salience, an additional salience signal was embedded in a standard ANN[17], such that it directly affected each individual unit during training, and collectively produced a reverse salience signal during testing (**Figure 1**). The input salience signal affects the sigmoidal threshold function of each node in the ANN by either reducing or increasing the threshold of the signal required to produce the original output signal (**Figure 2**). Similar ANN input combinations would be expected to produce similar reverse salience signals, because the reverse salience signal is defined as the summation of the nodal reverse salience signals observed at each node.

The threshold variations proposed in this research for use with the SANN closely mirror the variations described by Phil Husbands et al[14]. This research focuses primarily on the threshold offset variable, referred to as $b_i$ by Phil Husbands et al. [14], and which we incorporate in the definition of the threshold variable, which we refer to as $T_{new}$ in this research.

This research is different from the research on GasNets by Phil Husbands et al. as it models both the way memories associated with emotionally-laden events may be embodied in neural networks, and the way those emotional associations can be recalled when the events are remembered. Furthermore, this research introduces an additional input salience signal during training, and each node produces a nodal reverse salience signal during testing. This research also demonstrates that these kinds of networks enable one-time learning by associating strong salience signals with input combinations.

To test the SANN, a face recognition application was developed. In order for the computer software program to associate saliencies with the faces, both parts-based and whole-based feature-extraction algorithms were investigated, including *principal component analysis (PCA)*[18,19], *vector quantization (VQ)*[20,21,22] and *non-negative matrix factorization (NMF)*[20,23]. There is psychological and physiological evidence for parts-based representations in the brain. Parts-based representations emerge by virtue of two properties namely, the firing rates of neurons are never negative, and the synaptic strengths do not change sign[20]. For this research paper, NMF was selected as the optimal parts-based feature-extraction algorithm.

The NMF section of the source code was adapted from an existing NMF algorithm implementation[24]. Image data was provided by the CBCL face image database (http://cbcl.mit.edu/cbcl/).

For all the tests conducted, the SANNs were designed with 49 inputs, a single hidden layer of 10 units and a single output layer and trained with 200 iterations, using back-propagation. During the training of an ANN, each input combination needs to be associated with an output value. The criterion for defining output values is not significant, and therefore any criterion can be used. For the training of the SANN in this research, the output value was arbitrarily created from the images in the dataset, specifically the average pixel grayscale value for each image.

## General analysis and optimization

Having designed the SANN and implemented the SANN in software, a general analysis and thereafter variable optimization was required. Certain factors were observed and optimized, namely the residual reverse salience signal and the effects of the magnitude of the salience signal used to train the SANN. For the duration of the SANN analysis and optimization, the SANN was trained using the multiple-trial training technique.

### *Residual reverse salience signal value*

It was observed, due to the nature of the definition of reverse salience signal value, that a neural network, trained with salience, but tested with a "control input" still produced a non-zero reverse salience signal, which we will refer to as the residual reverse salience value. We define a "control input" as an input with all values set to 0.5. A correlation test was performed (**Figure 3**), and it was observed that this residual reverse salience signal value closely followed NN output values.

If a residual reverse salience signal did not exist, or if the residual reverse salience signal was not related to the output, then there would be a low correlation. From the test performed, a correlation coefficient of **0.985** was calculated. This indicates that a residual reverse salience signal exists, that is closely correlated to the output.

From this result, it is clear that the salience must be treated as a relative value (the current face relative to a previous value) rather than as an absolute value, to ensure that the value is useful.

### *Salience signal magnitude*

It was suspected that the introduction of an input salience signal would affect the training of the neural network. An experiment was performed to observe the effects on the learning curve of the neural network under multiple-trial training, for various magnitudes of salience (**Figure 4**).

It was observed that that the presence of an input salience signal retards the speed of response, deforming the shape of the learning curve. As expected, the fastest learning response of the NN occurs when the salience is set to 0. The presence of an input salience signal does not, however, affect the learning curve significantly.

## **Results**

To highlight the overall effect and significance of threshold variation in each node, multiple-trial training was investigated. Thereafter, single-trial training was simulated to test the hypothesis that a neural network could be trained with a single training iteration.

Multiple-Trial Learning Technique

As designed, we expect the variation of node thresholds to result in a varying reverse salience signal. As mentioned above, similar SANN input combinations would be expected to produce similar reverse salience signals, because the reverse salience signal is defined as the summation of the nodal reverse salience signals observed at each node. We can therefore predict that faces with similar features produce similar reverse salience signals. To test this hypothesis, an SANN was trained using a multiple-trial training method, and with a dataset of 200 unique images of faces. For the training data set used, certain images were similar. For example, images 2, 3, 9, 10 and 11 were each unique, but of the same person. Images 9, 10 and 11 were trained with an input salience signal of value 1, while the remaining images were trained with an input salience signal of value 0. The relative reverse salience signals were recorded for various faces (**Figure 5**).

To support the concept of the value system described by Edelman's and Beebe and Lachmann's theory of the three principles of salience, we desire the reverse salience signal to be independent of the SANN output signal (i.e. not significantly affecting the output signal of the neural network). To test this hypothesis, an SANN was trained using a multiple-trial training method, and with a dataset of 200 unique images of faces. An SANN was first trained with an input salience signal of value 0 applied to all input combinations, and then a second SANN was trained with an input salience signal of value 1 attached only to images 9, 10 and 11. Thereafter the output signals were recorded and compared, and correlation coefficients were calculated. A low correlation coefficient would indicate large variations in the output values, and a high correlation coefficient would indicate a small variation, which would be desired. A correlation coefficient of **0.9998** was calculated between the output sets, hence indicating negligibly small variations, as desired. This result indicates that the reverse salience signal can be regarded as significantly independent of, or conceptually orthogonal to the SANN output signal (i.e. with negligible variations).

Single-Trial Learning Technique

The embedded salience signals in an SANN could enable the training of neural networks with a single training iteration. To test this hypothesis, a NN was trained using a single-trial training method, and with a dataset of 200 unique images of faces. First the SANN was trained without the influence of input salience signals. Thereafter a single training iteration was executed, where an input salience signal of value 1 attached only to images 9, 10 and 11.

For the single-trial training methods, an amplification factor was introduced. The amplification factor corresponds to the magnitude of the salience (or significance) attached a certain input combination. Amplification factors in the integer range [1-6] were tested, to observe the effects of varying magnitude of the input salience signal, and the reverse salience signals were recorded for various faces. The results were compared to the multiple-trial learning techniques for the same dataset (**Figure 5**). A single-trial training method produced a similar reverse salience signal profile to the multiple-trial training method, although of a noticeably smaller magnitude. Correlation coefficients were calculated between the single-trial and the multiple-trial training, as a function of input salience signal amplification factors, and it was found that the higher the amplification factor (in the tested range), the smaller the variations between the profiles (**Figure 6**). This result indicates that an SANN can be trained with a single training iteration, with similar effects to multiple-trial training.

**Implications**

The research explored here has many implications, especially at the computational, biological and psychological levels.

At a computation level, the modeling of ongoing regulations in neural networks will allow us to extract more information (in the form of salience) from a standard ANN, without significantly adding to the complexity of the ANN structure. Furthermore, exploring the effect of heightened affective moments on a neural network will enable the future training of neural networks with a single training iteration.

At the biological level, an SANN enables modeling of the effects of the peculiar nature of synaptic connections as opposed to gap-junction connections. In synaptic connections the electric interneural signal is converted to a chemical signal that crosses the synaptic gap, and then gets converted back to an electrical signal. With the simplified model of gap-junction connections, the signal propagates as the direct transmission of an electric signal. The key point is that a synaptic connection allows non-local modulation of local synaptic processes via diffuse projection[1] of neurotransmitters to the synapse region, while the simplified model of gap-junction connections does not allow such effects. It is this non-local modulation of local synaptic processes that can be modeled by an SANN.

At a behavioural level, an SANN allows the modeling of the effect of affective states on brain activity and on memory (for example, adding an emotional tag to significant memories), because the ascending systems originate in the limbic system which is the seat of affective states. Thus a SANN potentially represents the effects of emotions on cortical activity, which are known to be significant[16].

**Conclusions**

This simple system displays the fundamental features and behavior that we associate with salience and learning.  It is implemented by a diffuse network of undifferentiated connections layered on a locally connected network of specific weights.  When presented with face images which have a high associated salience, the combined system not only learns rapidly to produce a strong output salience response, but also produces this response for other images which are very similar.  This salience processing operates independently to the basic face recognition processing performed by the locally-connected network.

Notably, when trained with single trials, the network produces similar reverse salience signal profiles to the multiple-trial training response. This indicates that an SANN can be trained with a single training iteration, with similar effects to multiple-trial training. It was found that the higher the input salience signal amplification factor (in the tested range), the closer single-trial training approximated multiple-trial training.  We associate this result with the expectation that single events of very high salience must produce a learned response; to the best of our knowledge, this is the first demonstration of this phenomenon in an artificial neural network.

**Future Work**

Future work includes, but is not limited to investigating alterative variations to the sigmoidal threshold function at each node in response to the input salience signal, varying the hidden layer size, and adapting the SANN to use a more biologically-accurate spiking-neuron NN.

**References**

1. Edelman, G. 2004. *Wider than the sky: A revolutionary view of consciousness.* USA: Yale University Press.
2. Bishop, C. M. 1999. *Neural Networks for Pattern Recognition*. Oxford: Oxford University Press
3. Hawkins, J. & Blakeslee, S. 2005. *On Intelligence*. Times Books: Henry Holt and Company, New York
4. Kingsley, R.RE.1996. *Concise Text of Neuroscience*. Lippincott Williams and Wilkins: Philadelphia.
5. Edelman, G. M., Tononi, G. 2000. *Consciousness: How matter becomes imagination*. Allen Lane, USA.
6. Edelman, G. M., Reeke, G., Gall, W., Tononi, G., Williams, D., Sporns, O. 1992. Synthetic neural modeling applied to a real-world artifact. *Proceedings of the National Academy of Sciences USA*. 89:7267-7271
7. Beebe, B. & Lachmann, F.M. 1994. Representation and Internalization in Infancy: Three Principles of Salience. *Psychoanalytic Psychology.* 11:127-165.
8. Rock, I., Heimer, W. 1959. Further Evidence of One-Trial Associative Learning. *American Journal of Psychology*. 72(1):1-16
9. Armstrong, C. M., DeVito, L. M. & Clelan, T. A. 2006. One-Trial Associative Odor Learning in Neonatal Mice. *Chem. Senses*. 31:343–349
10. Lee-Teng, e. & Sherman, m. 1966. Memory consolidation of one-trial learning in chicks. *Proc Natl Acad Sci USA*. 56(3):926–931
11. Waelti, P., Dickinson, A., Schultz, W. 2001. Dopamine responses comply with basic assumptions of formal learning theory. *Nature*. 412:43-48
12. Montague, P. R., Hyman, S. E. & Cohen, J. D. 2004. Computational roles for dopamine in behavioural control. *Nature*. 431:760-767
13. Downara, J., Mikulisa, D.J. & Davisa, K.D. 2003. Neural correlates of the prolonged salience of painful stimulation. *NeuroImage.* 20(3):1540-1551.
14. Husbands, P., Smith, T.M.C., O'Shea, M., Jakobi, N., Anderson, J. & Philippides, A. 1998. Brains, Gases and Robots. In Niklasson, N. et al. (editors). *Proceedings of the 8th International Conference on Artificial Neural Networks: ICANN98*. Springer.
15. P. Husbands. 1998. Evolving robot behaviours with diffusing gas networks. In P. Husbands and J.A. Meyer (editors): *Evolutionary Robotics*. 71–86. LNCS vol 1468, Springer-Verlag.
16. Damásio, A. 1995. *Descartes' error: Emotion, reason, and the human brain.* New York: Avon Books.

17. Arbib, M.A. 2003.The Elements of Brain Theory and Neural Networks. In *The Handbook of Brain Theory and Neural Networks*. Ed. M.A. Aribib. Second ed.Massachusatts: MIT Press. 1-24.
18. Dunteman, G.H. 1989. *Principal Components Analysis.* 2, illustrated ed.Newbury Park: SAGE.
19. Jolliffe, I.T. 2002. *Principal component analysis.* 2, illustrated ed. New York: Springer.
20. Lee, D. D. Seung, H. S. 1999. Learning the parts of objects by non-negative factorization. *Nature (London).* (6755):788-791.
21. Lee, D. D. Seung, H. S. 2001. Algorithms for Non-negative Matrix Factorization. *Advances In Neural Information Processing Systems.* (13):556-562.
22. Gersho, A. & Gray, R.M. 1992. *Vector quantization and signal compression.* Boston: Kluwer Academic Publishers.
23. Hoyer, P.O. 2004. Non-negative Matrix Factorization with Sparseness Constraints. *J.Mach. Learn. Res.* 5:1457-1469.
24. Hoyer, P.O. August 2006. *NMF Pack.* Helsinki, Finland: http://www.cs.helsinki.fi/.

**Figures**

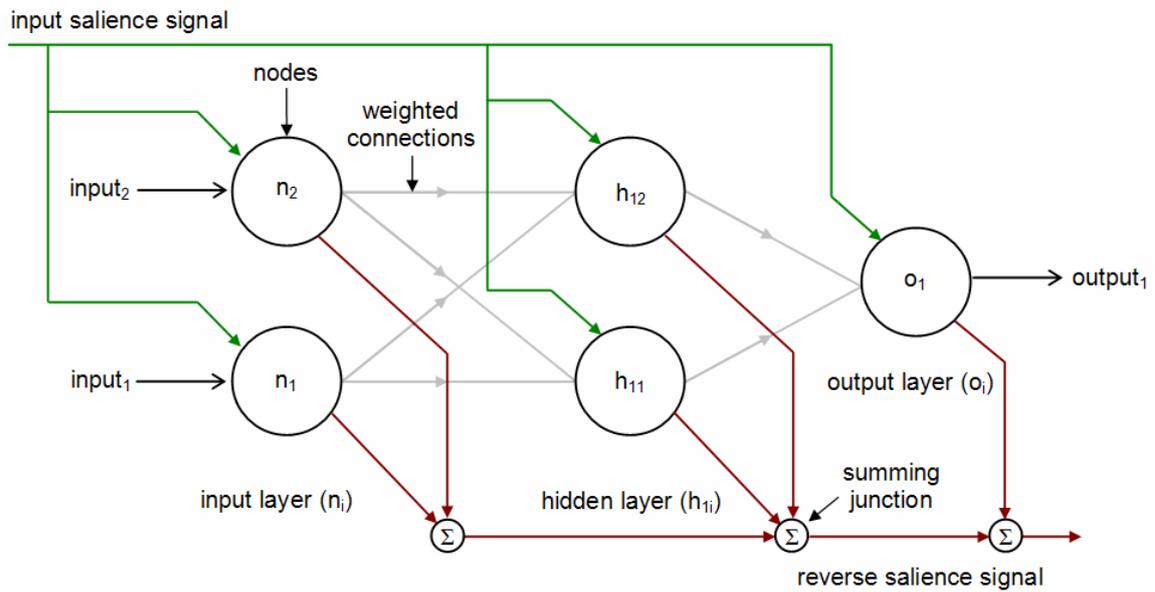

Figure 1: An SANN with a 2-node input layer ($n_i$), a 2-node hidden layer ($h_{1i}$), and a single output node ($o_i$). Each node receives an *input salience signal*, which has the effect of varying the threshold value of the sigmoidal function at each individual node, during training, depending on the node's activation level. From each node, *nodal reverse salience signals* are summed, to create a *reverse salience signal* for the SANN.

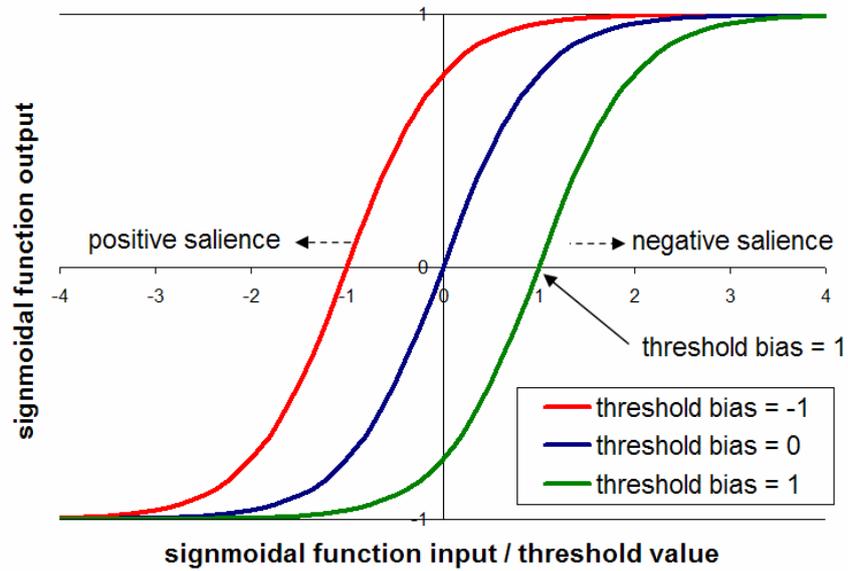

Figure 2: Varying thresholds of an approximate tanh-based sigmoidal function. It can be seen that the input salience signal applied to each node has the effect of pushing active node's threshold closer to their activation threshold (for positive salience), or moving them further away from this threshold (for negative salience).

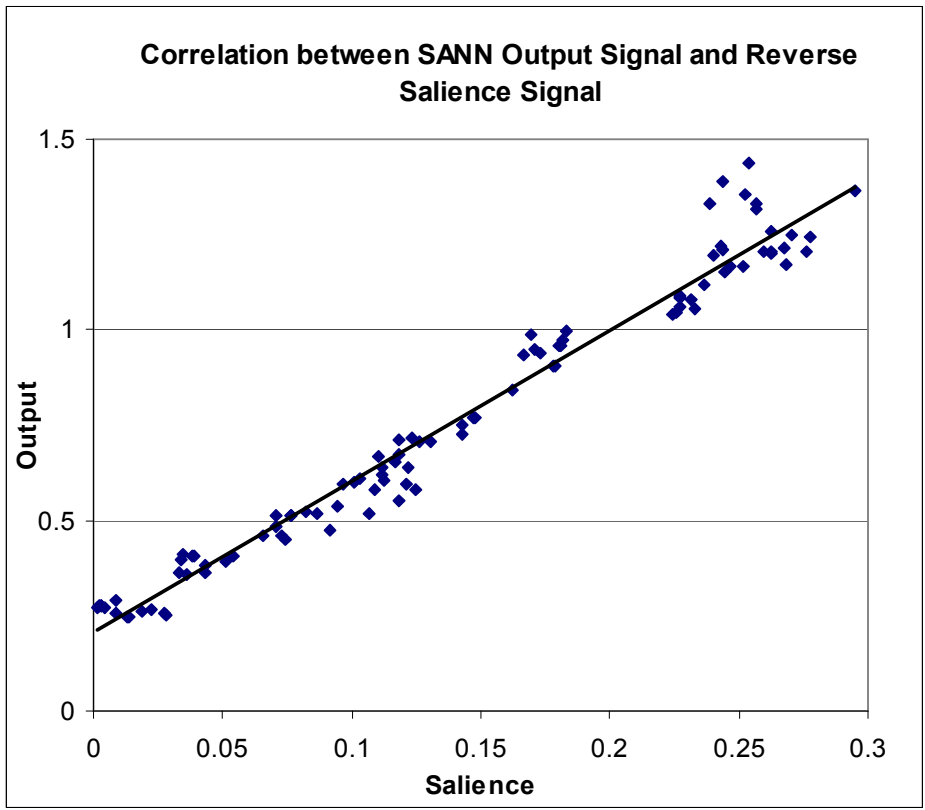

Figure 3: Correlation between SANN Output Signal and Reverse Salience signal after being trained and tested with 100 images from the CBCL database

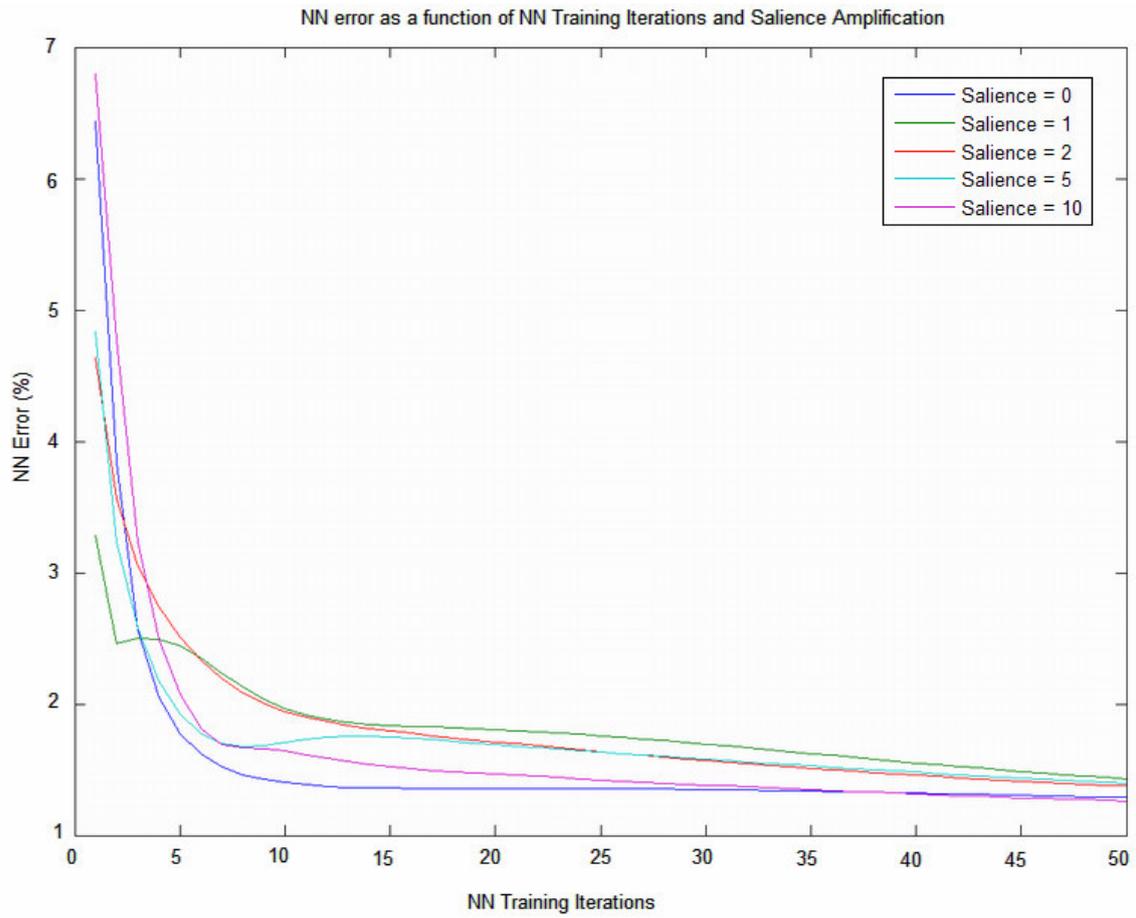

Figure 4: NN Error as a function of training iterations for various magnitudes of input salience signal values

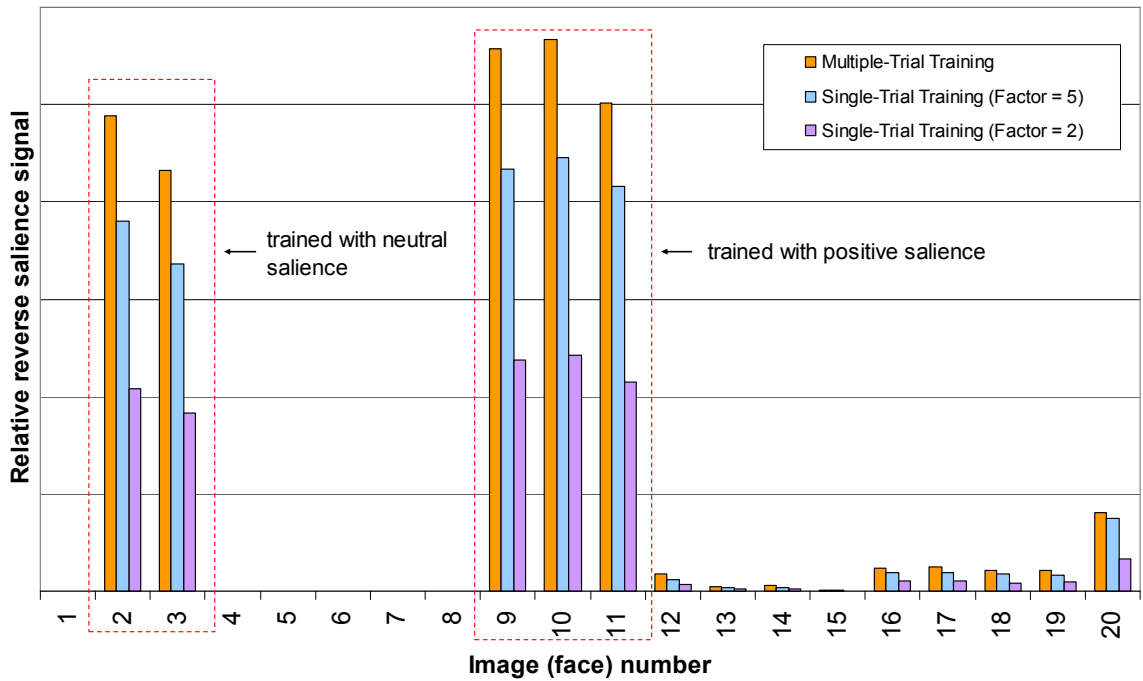

Figure 5: Relative reverse salience signals for various images of faces (a sample of 20 input combinations are shown). For the training data set used, certain images were similar. For example, images 2, 3, 9, 10 and 11 were each unique images, but were all photos of the same person, and therefore very similar. Images 9, 10 and 11 were trained with an input salience signal of value 1, while the remaining images were trained with an input salience signal of value 0. Noticeably, images 2 and 3 returned a relatively high reverse salience signal value, despite being trained with an input salience signal of value 0. For the single-trial training methods, an amplification factor was introduced. The amplification factor corresponds to the magnitude of the salience (or significance) attached a certain input combination. Amplification factors in the integer range [1-6] were tested, to observe the effects of varying magnitude of the input salience signal, and the reverse salience signals were recorded for various faces. Results attained from single-trial training with an amplification factors of 2 and 5 respectively, were plotted alongside multiple-trial training results, as seen above. A single-trial training method produced a similar reverse salience signal profile to the multiple-trial training method, although of a noticeably smaller magnitude.

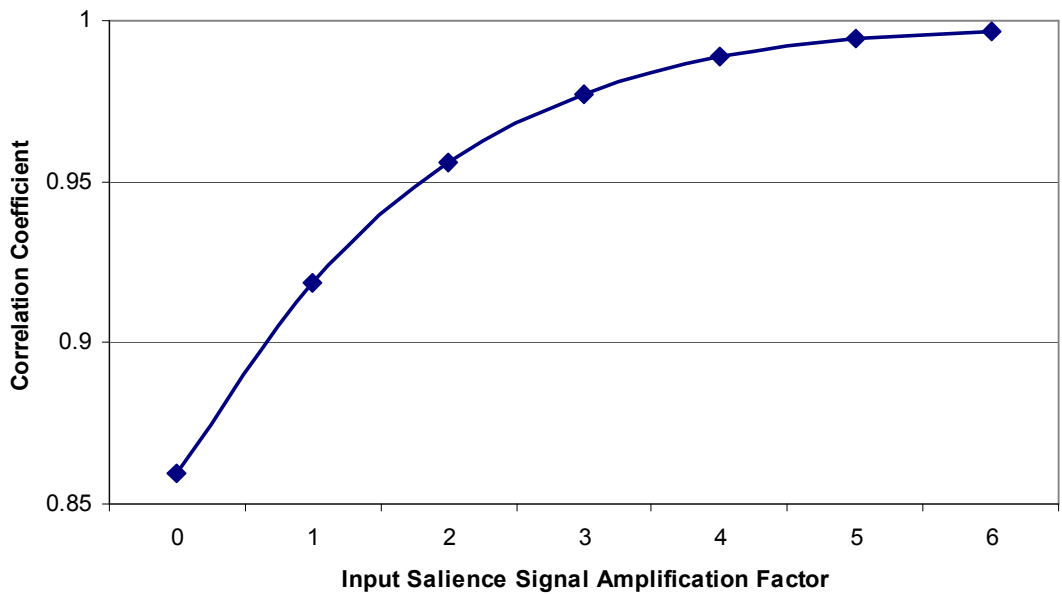

Figure 6: Correlation between single-trial and multiple trial reverse salience signal values, for varying input salience signal amplification factors during single-trial training. An SANN was trained using a single-trial training method, and with a dataset of 200 unique images of faces. First the SANN was trained without the influence of input salience signals. Thereafter a single training iteration was executed, where an input salience signal of value 1 attached only to images 9, 10 and 11. Amplification factors in the range [1, 2, 3, 4, 5, 6] were tested. Correlation coefficients were calculated between the single-trial and the multiple-trial training, as a function of input salience signal amplification factors. A high correlation coefficient would indicate a similar profile, which we desire.

# METHOD

## Salience signal

### *Threshold adjustment definition*

For the SANN, the *direction of threshold adjustment* was defined in terms of the salience signal S, and the current activation level $A_i$, as seen in (**Figure 7**). For example, if the node had a positive activation level and a positive salience level was assigned to it, the threshold should be reduced, to allow the node to produce a higher activation level the next time around. Conversely, increasing the threshold level will reduce the output signal of a node given any input signal.

|  | Salience + | Salience - |
|---|---|---|
| Activation + | -ve | +ve |
| Activation - | +ve | -ve |

**Figure 7: Direction of threshold adjustment as a function of salience and activation**

The adjustment factor ($D_{adj}$), in the integer range [-1, 0, 1], related to its current activation level $A_i$ and the magnitude of the input salience signal (S) is defined in **Equation 1**. The variable $U_{act}$ represents the current node activation.

$$D_{adj} = -\frac{U_{act} \times S}{|U_{act} \times S|} \qquad \textbf{Equation 1}$$

To prevent the threshold from increasing/decreasing indefinitely, *threshold limits* were defined. The salience-influence definition adjusts the threshold of the active nodes relative to the pre-defined threshold limits. The magnitude of the adjustment factor used in this research was 20% of the distance between the threshold and the threshold limit, in the appropriate direction. This method ensured that the threshold never exceeded the threshold limits.

The final definition of threshold variation for a node, with respect to its current activation level $A_i$ and its threshold limit $T_{limit}$ is defined as seen in **Equation 2**.

$$T_{new} = T_{old} - B \times \left(T_{old} + |U_{act}| \times D_{adj} \times T_{lim}\right)$$ **Equation 2**

In **Equation 2**, the variables $T_{new}$, $T_{old}$ and B represent the *new threshold*, *old threshold* and *salience influence* respectively. The *salience influence* is defined as the rate at which the thresholds were influenced by the salience signals.

*Reverse salience signal*

For this research, the *reverse salience signal* (S') experienced by active units was defined as the relationship between its current activation level $A_i$, the threshold $T_i$ and the sum of the weighted signals $V_i$ received by the units (**Equation 3**).

$$S' = A_i \times (T_i - V_i)$$ **Equation 3**

Although various methods could be used, a standard summing method was used in this research to collect the reverse salience signals from all the units, for simplicity. In the summing method, the individual reverse salience signals are summed, and the results are an indication of the salience attached to the various input-signal combinations.

*NMF Reconstruction algorithm*

The approximate reconstruction of the original data set can be attained by the **Equation 4**[23].

$$V_{i\mu} \approx (WH)_{i\mu} = \sum_{a=1}^{r} W_{ia} H_{a\mu}$$ **Equation 4**

Hoyer warned that while updating H (assuming W is fixed), one must be careful not to do any vector normalization in the iteration. Normalization of the rows of H makes sense when there are many columns, but not when there is a single column[24].

A NMF reconstruction algorithm was specifically designed for the purpose of this research, based on the **Equation 4** and recommendations provided by Hoyer[0].

## Software implementation

In order to implement the SANN application in software, a collection of data and certain source code was required.

### *Data collection*

Although the Matlab software application was designed to use real-time image capturing via the means of a webcam, additional tests were conducted using face images provided by the CBCL face image database. The face image database (CBCL data) can be found at: http://cbcl.mit.edu/cbcl/software-datasets/FaceData2.html

### *NMF source code*

The NMF section of the source code designed in this research was adapted from an existing NMF algorithm implementation[25]. As mentioned above, a NMF reconstruction algorithm was specifically designed for the purpose of this research, based on the equations and recommendations provided by Hoyer. The code for the NMF reconstruction algorithm was not provided in the original code written by Hoyer.

### *ANN source code*

The ANN section of the source code designed in this research was originally written in Python and placed in the public domain by Neil Schemenauer (nas@arctrix.com). For this research I adapted and translated the source code from Python into Matlab code.

The neural network is an artificial abstraction of the computational function of a biological neuron[26]. The program uses the standard backpropagation algorithm to train the MLP. The neural network used in this project has a single hidden layer of units.

*ANN input, hidden and output layers*

For this research, the neural network was designed with 49 inputs, a single hidden layer of 10 units and a single output layer. The inputs used for the neural network were the 49 weights assigned to a different element in the mixing matrix (W). For a neural network to operate and be trained an output value must be chosen. For this research, an arbitrary, easy-to-calculate output variable was created from the images in the dataset, namely the *average* pixel grayscale value for each image.

*ANN training and testing*

For this research, the neural network was trained with 200 iterations for all tests performed.

## General analysis and optimization

*Introduction*

Having designed the SANN and implemented the SANN in software, a general analysis and thereafter variable optimization was required. Certain factors, including the number of units in the single hidden layer of the SANN, were optimized. For the duration of the SANN analysis and optimization, the SANN was trained using the multiple-trial training technique.

*Hidden layer units*

It is known that that too few hidden layer nodes will not allow for efficient training of the NN, and initially increasing the hidden layer number will allow for the network to be trained with less error. However, further increasing the size of the hidden layer should result in the over training of the NN, at which point the NN will no longer be able to generalize, which will result in an increase in overall NN error.

The number of units in the single hidden layer of the SANN was varied to observe the effect on the learning curve, using a salience amplification of 2. The test was performed on a range of hidden layer sizes [2:18], and the number of iterations for the NN to show a 90% improvement in the error was recorded. The smaller the number of iterations, the easier the NN is to train (**Figure 8**).

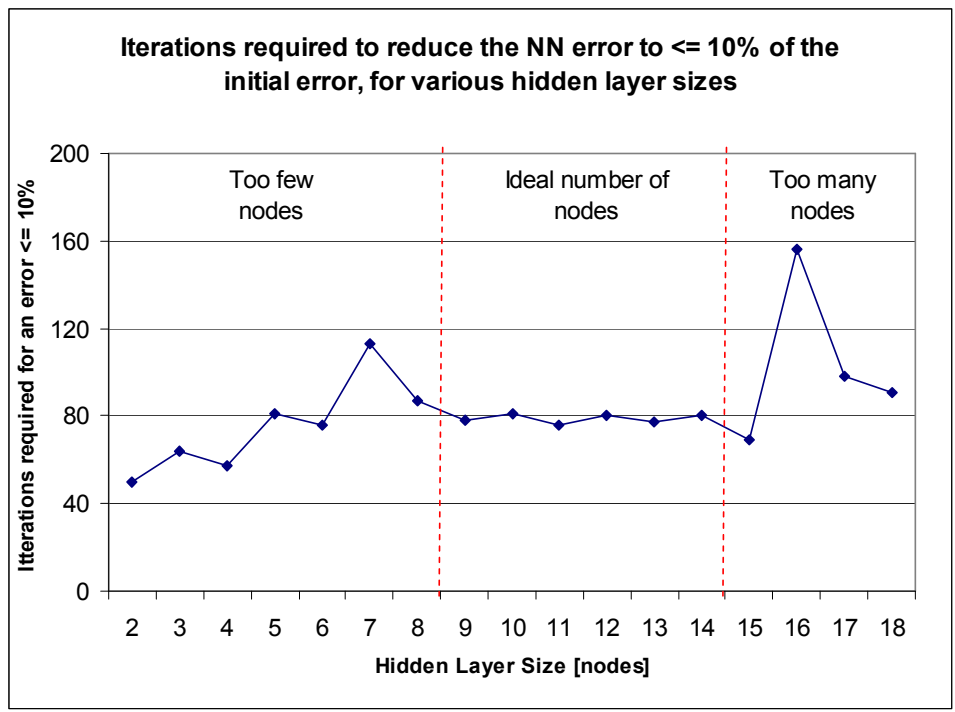

**Figure 8: Iterations required to reduce the NN error to <= 10% of the initial error, for various hidden layer sizes.**

As expected, as the size of the hidden layer increased, the time required to train the NN to within 10% error decreased. A hidden layer less than 9 was considered too small to effectively train, and a hidden layer size greater than 14 was considered too large (becomes too specifically trained). The size selected for a hidden layer in this research was between 10 and 13 nodes.

# References

23. Hoyer, P.O. 2004. Non-negative Matrix Factorization with Sparseness Constraints. *J.Mach. Learn. Res.* 5:1457-1469.

24. Hoyer, P.O. & Remmelzwaal, L.A. 2009. *nmfpack Question.* RSA: Email.

25. Hoyer, P.O. August 2006. *NMF Pack.* Helsinki, Finland: http://www.cs.helsinki.fi/.

26. Maass, W. 2003. Spiking Neurons, Computation with. *The Handbook of Brain Theory and Neural Networks.* :1080-1083.